\documentclass{article}
\usepackage{spconf,amsmath,graphicx,amssymb,algorithm}
\usepackage{algcompatible}
\usepackage{color}

\usepackage[spaces,hyphens]{url}
\usepackage{enumitem}
\usepackage{algpseudocode}
\algrenewcommand\algorithmicrequire{\textbf{Input:}}
\algrenewcommand\algorithmicensure{\textbf{Output:}}

\title{PRIORITY-BASED POST-PROCESSING BIAS MITIGATION FOR INDIVIDUAL AND GROUP FAIRNESS}
\name{\begin{tabular}{c}$^{\dagger}$Pranay Lohia\end{tabular}}
\address{$^\dagger$IBM Research AI, India}

\newtheorem{definition}{Definition}
\begin{document}
\ninept
\maketitle
\begin{abstract}
Previous post-processing bias mitigation algorithms on both group and individual fairness don't work on regression models and datasets with multi-class numerical labels. 
We propose a priority-based post-processing bias mitigation on both group and individual fairness with the notion that similar individuals should get similar outcomes irrespective of socio-economic factors and more the unfairness, more the injustice.
We establish this proposition by a case study on tariff allotment in a smart grid. Our novel framework establishes it by using a user segmentation algorithm to capture the consumption strategy better. This process ensures priority-based fair pricing for group and individual facing the maximum injustice. It upholds the notion of fair tariff allotment to the entire population taken into consideration without modifying the in-built process for tariff calculation. 
We also validate our method and show superior performance to previous work on a real-world dataset in criminal sentencing.
\end{abstract}
\section{Introduction}
\label{sec:intro}

Over the past few years, there has been a rapid development in the domain of machine learning and its application. As a consequence of that, machine learning systems are now being used as a tool to make high-stakes decisions, which could affect an individual's daily life. However, it has been observed that these machine learning systems sometimes produce an outcome that is unfair or discriminatory for minorities, historically disadvantaged populations, and other groups. For example, in COMPAS \cite{dressel2018accuracy}, the algorithm used for recidivism prediction produces a much higher false-positive rate for black people than white people, therefore being discriminatory for black people.

Bias creeps into the machine learning model because of many reasons. Skewed training sample, sample size disparity, and proxies are a few of the many reasons which lead the system to be unfair towards one group. In general, the assumption of one size fits all policy lead to discrimination among groups. There are two central notions of fairness in decision making: individual fairness and group fairness
\cite{8682620}. While individual fairness, in a broader sense, requires similar individuals should be treated similarly whereas group fairness seeks for some statistical measure to be equal among group defined by protected attributes (such as age, gender, race, and religion). Disparate impact (DI) is a standard measure for group fairness.

Pre-processing, in-processing, and post-processing are three stages for performing debiasing in the machine learning model. Pre-processing involves transforming the feature space into another feature space which, as a whole, is independent of the protected attribute. The in-training algorithm adds a regularization term in the objective function of the model to achieve independence from a protected attribute. While in the post-processing algorithm, the goal is to edit the outcome to achieve fairness.

Several works have been done on price fairness in a smart grid. In \cite{maestre2018reinforcement}, the author has used Q-Learning with neural network approximation to assign the right price to the individual of different groups. In \cite{han2014impact}, the author shows a strong correlation between the socio-economic factor and the load consumption pattern.  In this paper, we first propose a tariff allotment policy. The electricity load forecasting task is a time-series forecasting problem. SVR has been found to perform better for time-series forecasting \cite{muller1997predicting}. We use SVR to predict day-ahead aggregated load for each of the consumers for the entire day. The reason for day-ahead allocation being that it makes the consumer aware of the power consumption. The power market prefers tariff slab or bucketing for pricing \cite{sarker2014smart}. For creating slabs, we use the Gaussian bucketing method to bin the predicted value and assign respective tariff to individuals.  

An application-independent priority-based post-processing debiasing algorithm with the notion of both group and individual fairness is used to remove the bias associated with the previous step that uses machine learning as an architecture. In most real scenarios, we won't have access to the internals of models. Therefore it is evident that the post-processing debiasing algorithm with the notion of both group and individual fairness suits the most \cite{8682620}. The trade-off between Bias and Accuracy \cite{feldman2015certifying} limits the number of individual samples allowed to be debiased. This brings the requirement of individuals facing more unfairness to be debiased in priority \cite{berk2017convex}. Using this approach, we can achieve fairness faster both in time and number of samples whose label needs to be changed to achieve fairness. To the best of our knowledge, all existing post-processing algorithms don't work on priority-based debiasing \cite{8682620,KamiranKZ2012,HardtPS2016,PleissRWKW2017,CanettiCDRSS2018}. The starting point for our proposed approach is the individual bias detector of \cite{aggarwal2019black}, which finds samples whose model prediction changes when only protected attributes change and calculate \emph{Unfairness Quotient}, which is used to identify the priority order of the samples.

We have two major contributions to this work: 
\begin{enumerate}
    \item An application-independent priority-based post-processing debiasing algorithm that improves both individual and group fairness.
    \item A case study to showcase a fair tariff allocation strategy which charges a similar price to an individual with similar load characteristics irrespective of socio-economic factors.
\end{enumerate}

Compared to the randomized debiasing method used in \cite{8682620}, we have superior performance. We validate our proposition on a real-world dataset in criminal sentencing \cite{sentencing}. The remainder of the paper is organized as follows. First, we provide details on the tariff allocation policy using \cite{ISSDA} in Sec.3. Next, we propose a priority-based post-processing debiasing algorithm in Sec.4 with results on real-world datasets \cite{ISSDA} \cite{sentencing} in Sec.5. We have showcased the comparison result with \cite{8682620}. Finally, we conclude the paper in Sec.6.

\clearpage

\section{FAIRNESS}
\label{sec:fairness}
 Before proposing a priority-based post-processing debiasing algorithm, we first introduce the working definitions of two important notions of fairness i.e. group and individual fairness. Consider a supervised classification problem with features $\mathbf{X} \in \mathcal{X}$, protected attributes $\mathbf{P} \in \mathcal{P}$, and labels $\mathbf{C} \in \mathcal{C}$.  We are given a set of training samples $\{(x_1,p_1,c_1), \ldots, (x_n,p_n,c_n)\}$ and would like to learn a classifier $\hat{C}: \mathcal{X} \times \mathcal{P} \rightarrow \mathcal{C}$. For ease of explanation, we will only consider a scalar binary protected attribute, i.e.\ $\mathcal{P} = \{0,1\}$, and a binary classification problem, i.e.\ $\mathcal{C} = \{0,1\}$. The value $p = 1$ is set to correspond to the \emph{privileged / majority} group (e.g.\ male as in this application) and $p = 0$ to \emph{unprivileged / minority} group (e.g.\ female).  The value $c = 1$ is set to correspond to a \emph{favorable} outcome (e.g.\ receiving a loan or not being arrested) and $c = 0$ to an \emph{unfavorable} outcome.

\subsection{Group Fairness}

\begin{definition}

We define group fairness in terms of disparate impact \cite{feldman2015certifying} as follows.  There is disparate impact if
\begin{equation}
    \label{eqn:disp_imp}
    \frac{\mathbb{E}[\hat{c}(\mathbf{X},P) \mid P = 0]}{\mathbb{E}[\hat{c}(\mathbf{X},P) \mid P = 1]}
\end{equation}
is less than $1 - \epsilon$ or greater than $(1 - \epsilon)^{-1}$, where a common value of $\epsilon$ is 0.1 or 0.2. In our method, the value of $\epsilon$ is 0.1.

\end{definition}

In our designed scheme, group fairness implies that the prediction for an individual across different groups should be almost equiprobable.

\subsection{Individual Fairness}
\begin{definition}
From the data, a given sample $i$ has individual bias if $\hat{c}(x_i,p=0) \neq \hat{c}(x_i,p=1)$.
\end{definition}

In our designed scheme, individual fairness confirms that similar individuals should be treated similarly i.e. prediction allotted to a particular individual should not be sensitive to the protected attributes of the individual \cite{8682620}.

\section{Case Study: TARIFF ALLOTMENT}
\label{sec:case}

\subsection{Dataset:}
For the evaluation purpose, The CER ISSDA dataset \cite{ISSDA} is used which was obtained by the Irish Commission for Energy Regulation (CER) for the purpose to analyze the impact on consumer's electricity consumption. It contains half-hourly load consumption data points of around 5000 Irish households and SMEs. In addition to the load data, the dataset also includes some survey questionnaire which describes the socio-economic condition of the consumer.

\subsection{Learning Stage:}
The inputs features are past load values, temperature, calendar details, and socio-economic factors. Our day-ahead load forecasting algorithm uses Adaboost regressor \cite{scikit-learn} with the base learner as support vector regression (SVR). Adaboost regressor is a meta-estimator that begins by fitting the base estimator which in this case is SVR. We use Root mean squared error (RMSE) calculation to evaluate the model (Adaboost+SVR) and found to be better performing with a value of 8.3 against 10.54 in the base estimator case.

\subsection{Allotment Policy:}
We are proposing a day-ahead tariff allocation policy to assign a tariff band or slab to the individual \cite{sarker2014smart}. The role of tariff bands is that all the individuals who are assigned to the same tariff band are charged the same amount. To distinguish the consumer, we use the Gaussian bucketing technique. The reason behind choosing this method to represent tariff bands was that the probability distribution of the dataset is Gaussian in nature. Assume the prediction from model be ${\hat y_1,\hat y_2,\dots,\hat y_n}$, we calculate the mean $\mu$ and the standard deviation $\rho$ and assign all the individual same tariff whose predicted value lies in $\left[\mu-i\rho,\mu-(i-1)\rho \right ) \text{\ or \ } [\mu+i\rho,\mu+(i+1)\rho) \ \ \forall i \in \mathbb{Z}$. The algorithm \ref{alg:ALG1} is used to assign tariff band. The input to algorithm is predicted load ($\hat{y}$) while output is tagged tariff band ($tid$). The time complexity of the algorithm to assign tariff to each individual is $\mathcal{O}(n)$ where $n$ is the number of individual in the system.

\begin{algorithm}
\caption{Tariff allotment policy }
\label{alg:ALG1}
\begin{algorithmic}
\Require $\tau$ = $\{\langle id_1,\hat y_1 \rangle,\langle id_2,\hat y_2\rangle,\dots,\langle id_n,\hat y_n\rangle\} \subset ID \times \hat Y $
\Ensure $\{\langle id_1,tid_1 \rangle,\langle id_2, tid_2\rangle,\dots,\langle id_n, tid_n\rangle\} $
\Function {AssignTariff}{$\tau$}
\State tariffAllotlist $\gets \{\ \}$
\State {$\mu$,\ \ $\rho$ = mean$(\tau)$,\ \ \text{stdDev}$\left (\tau\right )$} \Comment{mean and stdDev of $\hat y \in \tau$}
\ForAll  {$\langle id,\hat y \rangle \in \tau$}
\State tariffAllotlist.append($\langle id, \frac{x-\mu}{\rho}\rangle$)
\EndFor 
\State leastTariff = min(tariffAllotlist) \Comment{return min value of second element of tuples}
\State \Return tariffAllotlist
\EndFunction
\end{algorithmic}
\end{algorithm}

\section{PRIORITY-BASED POST-PROCESSING BIAS MITIGATION}
\label{sec:debiasing}
The policy described in the previous section doesn't take into account the biased behavior of the model concerning the protected attributes known at runtime i.e. impact of socio-economic factors. Before we start discussing priority-based post-processing methodology for removing group and individual biases concerning the protected attribute (age, race, and gender), one may argue that the debiasing phase would not be required if during model training we would have removed these sensitive attributes as features. However, this method poses this weakness because the non-sensitive attribute can behave as a proxy for the sensitive attribute that may not be visible at first \cite{kusner2017counterfactual}. Therefore a post-processing debiasing scheme is proposed. Another advantage which post-processing debiasing process provides is that it can be applied to any black-box model.

 Bias and Accuracy always have a trade-off \cite{feldman2015certifying}. Post-processing individual bias mitigation may result in drop in accuracy. While performing the debiasing process, we always have to consider a bandwidth of accuracy loss allowed and thereby limiting the number of individuals allowed to be debiased. Hence, we are using group discrimination metric (Disparate Impact) as a threshold to limit the number of individual samples to be debiased. We terminate the debiasing process once the fair threshold is reached or it gets terminated by itself \cite{8682620}. We define \emph{Unfairness Quotient} as the difference between the actual model prediction and the prediction after perturbing. It signifies the amount of unfairness associated.
 Priority-based debiasing is needed to ensure priority-based justice to the individual facing maximum unfairness. This allows individuals facing more bias to be debiased in priority. This algorithm can be implemented on N-nary and continuous-valued protected attributes (shown in Figure \ref{tariffAgeDi}) since they can always be categorized into privileged and unprivileged i.e. binary situation.

\subsection{Individual Bias detection and Unfairness Scoring}
\label{ssec:individualbias}

Consider the model $\hat c$ which comprises prediction followed by bucketing. To determine whether the sample $(x_i,d_i)$, where $d_i$ is the sensitive attributes and $x_i$ are the non-sensitive attributes, has individual bias associated with it, we perturb the sensitive attribute $d_i$ to $\check{d_i}$. If $\hat c(x_i,d_i)$ is different from  $\hat c(x_i,\check{d_i})$, we add this sample to individual biased sample collection. 

\begin{definition}
For each sample $(x_i,d_i)$, we calculate Unfairness Quotient $b_{x_i,d_i}$. Formally, Unfairness Quotient is defined as the difference between the actual model prediction and the prediction after perturbing.i.e
\begin{align}
    b_{x_i,d_i} = abs(\hat c(x_i,\check{d_i}) - \hat c(x_i,d_i)) 
\end{align}

\end{definition}

The \emph{Unfairness Quotient} signifies the amount of bias associated with that sample, i.e. more the value, more the injustice and hence higher the priority during debiasing.

\subsection{Overall Algorithm}
\label{ssec:algorithm}

The priority-based bias mitigation algorithm is applied after the model prediction and band allotment for each sample. For each sample, in the decreasing order of \emph{Unfairness Quotient}, we update the value of $\hat c(x_i,d_i)$ to  $\hat c(x_i,\check{d_i})$ until the value of DI is less than $1-\epsilon$. The proposed algorithm is summarized in Algorithm \ref{alg:ALG2}.

\begin{algorithm}
\caption{Priority-based post-processing bias mitigation algorithm}

\label{alg:ALG2}
\begin{algorithmic}
\Require Given model $\hat{c}$ trained on training set $\{(\mathbf{x}_i,d_i,y_i)\}$ where ($d_i$ is sensitive attribute and Test Sample $(\mathbf{X},\mathbf{D})$.)
\State{minority = 0}
\State{$di = \frac{\mathbb{E}[\hat{c}(\mathbf{X},\mathbf{D}) \mid d = minority ]}{\mathbb{E}[\hat{c}(\mathbf{X},d) \mid d = 1\oplus minority]} $}
\If {di $>$ 1}
\State minority = 1
\State {di = $\frac{1}{di}$}
\EndIf

\ForAll  {run-time test samples  $(\mathbf{x}_k,d_k)$}
\State { $b_{\mathbf{x}_k,d_k} = $ abs($\hat{c}(\mathbf{x}_i,d_i=0) - \hat{c}(\mathbf{x}_i,d_i=1)$)} \Comment{compute Unfairness Quotient}
\EndFor 
\State{Sort test-sample $(\mathbf{X},\mathbf{D})$ on the basis of $b_{\mathbf{x}_k,d_k}$ in decreasing order }
\ForAll{($\mathbf{x}_k,d_k) \in test-sample$ $(\mathbf{X},\mathbf{D})$}
\If{$b_{\mathbf{x}_k,d_k} > 0 \ \&\  d_k = minority$}
\State {$\hat{c}(\mathbf{x}_k,d_k) = \hat{c}(\mathbf{x}_k,d_k= 1\oplus minority )$}
\State{$di = \frac{\mathbb{E}[\hat{c}(\mathbf{X},\mathbf{D}) \mid d = minority ]}{\mathbb{E}[\hat{c}(\mathbf{X},d) \mid d = 1\oplus minority]} $}
\EndIf
\EndFor 
\end{algorithmic}
\end{algorithm}


\section{Empirical Results}
\begin{figure}[!htbp]
\centering
\includegraphics[height=2.25in,width=2.25in]{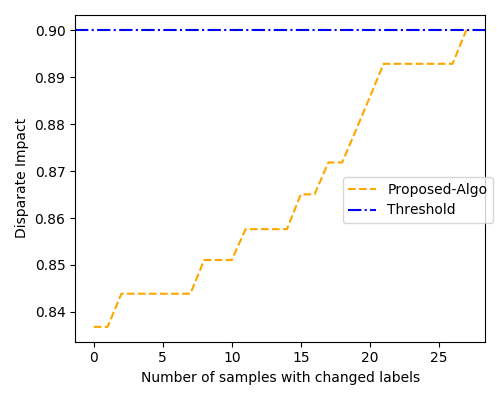}
\caption{Variation of Disparate Impact value, with the protected attribute as \textbf{Age} in the \textbf{Tariff dataset}, as the class label of individual samples is changed (the individual samples are debiased) \textbf{priority-based}. The debiasing process stops once the purple threshold line of 0.9 is reached.}
\label{tariffAgeDi}
\end{figure}

We evaluated our proposed algorithm on the subset of the CER ISSDA dataset described in subsection 3.1 for Jan 2010. We started by training an Adaboost + SVR model on the first 20 days sample to predict the power consumption by the individual followed by the Gaussian bucketing for tariff band allocation. We empirically verified the distribution between average power consumption and the number of individuals and found it to be Gaussian with mean 32.24 and variance 94.67. The training samples contain two protected attributes i.e. \emph{Gender} and \emph{Age} concerning which model can have group as well as individual unfairness. Since our proposed priority-based post-processing algorithm is application-independent, we verified the effectiveness of the algorithm on the tariff allotment model. The notion of fairness requires the definitions of group, i.e. privileged and unprivileged and output class labels, i.e. favorable and unfavorable. For the \emph{Gender} attribute, the group comprises Male and Female, and for the \emph{Age} attribute, we define the group as young i.e with Age $\leq$ 45 and old i.e Age $>$ 45. We used the percentage strategy and chose lower 40\% as favorable and upper 60\% as unfavorable which makes sense i.e. one is in a favorable zone if he/she is paying less than average. Figure \ref{tariffAgeDi} showcases the result of priority-based algorithm for protected attribute \emph{Age}.

\label{sec:result}

\subsection{Comparison with the baseline approach}
\begin{figure}[!htbp]
\centering
\includegraphics[height=3.25in,width=3.25in]{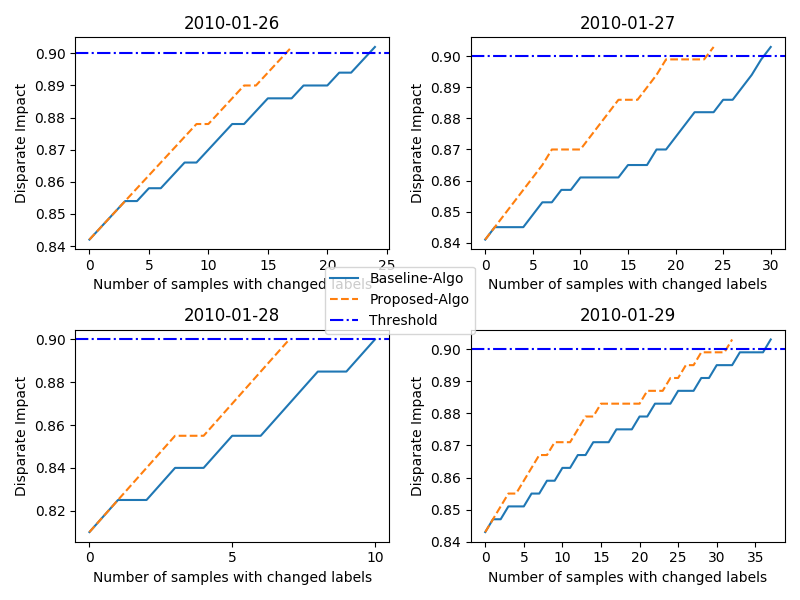}
\caption{Variation of Disparate Impact value, with the protected attribute as \textbf{Gender} in the \textbf{Tariff dataset} for multiple dates, as the class label of individual samples is changed (the individual samples are debiased). The orange line shows the variation according to the \textbf{priority-based} algorithm. The blue line shows the variation according to the baseline-randomized approach. The number of samples whose labels need to change to achieve fairness is less in our \textbf{priority-based} approach. The debiasing process stops once the purple threshold line of 0.9 is reached.}
\label{tariffGenderDi}
\end{figure}

We compared our priority-based debiasing approach with the baseline method in {\cite{8682620}}. In our case, daily in the post-processing stage, the model assigns tariff to each individual. We start by calculating the initial Disparate Impact which is a measure of group fairness discussed in Section 2.1 and if it comes out to be less than 0.9 we uplift its value towards 0.9, if achievable by debiasing the sample with individual bias. Due to the efficiency in selecting the individual bias point i.e in decreasing order of \emph{Unfairness Quotient} rather than randomly in {\cite{8682620}}, we found out that our approach takes less time as well as a lesser number of class label changes takes place compared to the existing algorithm. Due to the fewer number of changes in class label, our algorithm kind of reduces the unescapable bias-accuracy tradeoff. Figure \ref{tariffGenderDi} shows the comparison plot for both the algorithm with the protected attribute as \emph{Gender}.

\begin{figure}[!htbp]
\centering
\includegraphics[height=2.5in,width=2.5in]{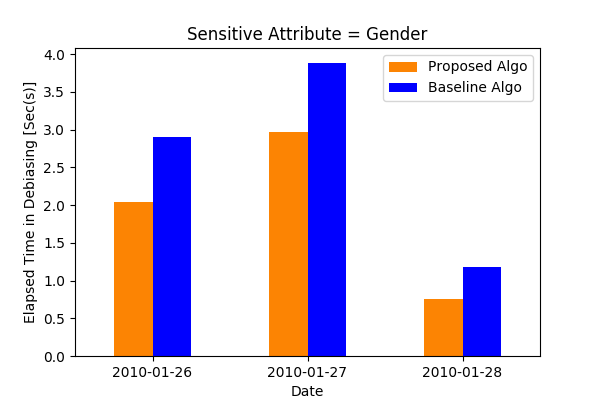}
\caption{Time taken by the proposed and the baseline approach in the \textbf{Tariff dataset} with the protected attribute as \textbf{Gender} over multiple dates. From all the bar plots, time taken is \textbf{priority-based} approach is lesser.}
\label{tariffGenderTime}
\end{figure}

\subsection{Validation of the proposed algorithm}
\begin{figure}[!htbp]
\centering
\includegraphics[height=2.25in,width=2.25in]{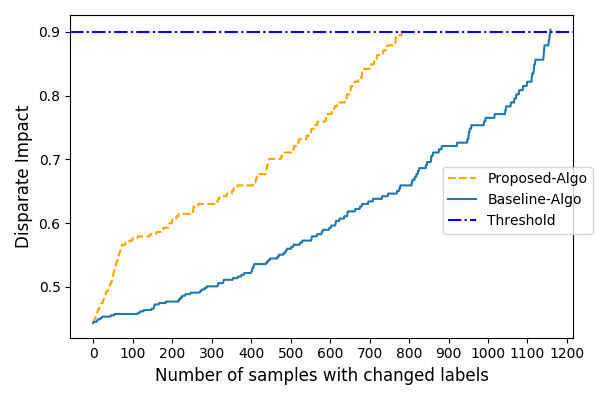}
\caption{Variation of Disparate Impact value, with the protected attribute as \textbf{Gender} in the \textbf{Sentencing dataset}, as the class label of individual samples is changed (the individual samples are debiased). The orange line shows the variation according to the \textbf{priority-based} algorithm. The blue line shows the variation according to the baseline-randomized approach. The number of samples whose labels need to change to achieve fairness is less in our \textbf{priority-based} approach.}
\label{SentenceGender}
\end{figure}

We extended our analysis to see how our algorithm performs on other publicly available datasets with multi-class numerical labels. For the validation purpose, we run the algorithm on Sentencing dataset \cite{sentencing}, which has \emph{Gender} as a protected attribute. On studying the dataset, we found out the charge imposed on Male was much higher than Female leading to some unfairness towards Male. Using our proposed algorithm, we were able to remove this unfairness by changing 784 class labels as compared to the 1163 class labels in the case of baseline. Also, the time taken by the debiasing process using the priority-based approach was 125s whereas using the baseline randomized approach was 190s on average. Figure \ref{SentenceGender} showcases this analysis.

\vspace{-1cm}
\section{CONCLUSION }
\label{sec:conclusion}

Algorithmic fairness has become a compulsory step to be taken into consideration when building and deploying machine learning models. Bias-mitigation algorithms that address the critical and important notion of fairness are important. Accuracy and Bias are always in a trade-off. This creates a limit on the number of individual samples to be debiased withholding the accuracy. In this constraint individual debiasing process, past randomization-based debiasing approach fails to give a validity on whether individual samples facing maximum unfairness have been taken into consideration for debiasing or not. In this paper, we have developed a novel priority-based post-processing algorithm that ensures priority-based fairness to the individual facing maximum unfairness. The remediation process work to improve both individual and group fairness metrics. In this shifting paradigms in the model development process, there is limited accessibility for deployers to the internals of trained models. Therefore, post-processing algorithms that treat models as complete black-box are necessary. In comparison to previous work, our proposed algorithm is faster and needs changing labels of a lesser number of samples to achieve fairness. We have validated our method on multiple real-world datasets. 

\clearpage



\bibliographystyle{IEEEbib}
\bibliography{template}

\end{document}